\documentclass[letterpaper, 10 pt, conference]{ieeeconf}  

\IEEEoverridecommandlockouts                              

\usepackage{tabularx}
\usepackage{amsmath,amsfonts}
\usepackage{xcolor}
\usepackage{threeparttable}
\usepackage{cite}
\usepackage{multirow}
\usepackage{longtable}
\usepackage{graphicx}
\usepackage{caption}
\usepackage{subcaption}
\usepackage{float}

\usepackage{soul}
\usepackage{url}
\usepackage{booktabs}
\usepackage{algorithm}
\usepackage{algorithmic}
\usepackage[switch]{lineno}

\usepackage{lineno}
\usepackage[hidelinks]{hyperref}

\usepackage{multirow}
\usepackage{threeparttable}

\hypersetup
{
 unicode=false,     
 pdftoolbar=true,    
 pdfmenubar=true,    
 pdffitwindow=false,   
 pdfstartview={FitH},  
 pdftitle={My title},  
 pdfauthor={Author},   
 pdfsubject={Subject},  
 pdfcreator={Creator},  
 pdfproducer={Producer}, 
 pdfkeywords={keywords}, 
 pdfnewwindow=true,   
 colorlinks=true,    
 linkcolor=red,     
 citecolor=blue,    
 filecolor=magenta,   
 urlcolor=cyan      
}

\title{\LARGE \bf
HeCoFuse: Cross-Modal Complementary V2X Cooperative Perception with Heterogeneous Sensors
}

\author{Chuheng~Wei$^{*}$, ~\IEEEmembership{Member,~IEEE},
        Ziye Qin,~\IEEEmembership{Member,~IEEE}, 
        Walter Zimmer,~\IEEEmembership{Member,~IEEE},\\
        Guoyuan~Wu,~\IEEEmembership{Senior~Member,~IEEE} 
        and Matthew~J.~Barth,~\IEEEmembership{Fellow,~IEEE},
\thanks{Chuheng Wei, Ziye Qin, Guoyuan Wu, and Matthew J. Barth are with the College of Engineering, Center for Environmental Research and Technology, University of California at Riverside, Riverside, CA, 92507 USA. }
\thanks{Ziye Qin is also with the School of Transportation and Logistics, Southwest Jiaotong University, Chengdu, China.}
\thanks{Walter Zimmer is with the Chair of Robotics, Artificial Intelligence and Real-time Systems, TUM School of Computation, Information and Technology, Technical University of Munich, 85748 Munich, Germany.}
\thanks{$^{*}$Corresponding author. e-mail: chuheng.wei@email.ucr.edu}
}

\begin{document}

\maketitle
\begin{abstract}
Real-world Vehicle-to-Everything (V2X) cooperative perception systems often operate under heterogeneous sensor configurations due to cost constraints and deployment variability across vehicles and infrastructure. This heterogeneity poses significant challenges for feature fusion and perception reliability. To address these issues, we propose HeCoFuse, a unified framework designed to enable effective cooperative perception across diverse sensor combinations, a unified framework designed for cooperative perception across mixed sensor setups—where nodes may carry Cameras (C), LiDARs (L), or both. By introducing a hierarchical fusion mechanism that adaptively weights features through a combination of channel-wise and spatial attention, HeCoFuse can tackle critical challenges such as cross-modality feature misalignment and imbalanced representation quality. In addition, an adaptive spatial resolution adjustment module is employed to balance computational cost and fusion effectiveness. To enhance robustness across different configurations, we further implement a cooperative learning strategy that dynamically adjusts fusion type based on available modalities.
Experiments on the real-world TUMTraf-V2X dataset demonstrate that HeCoFuse achieves 43.22\% 3D mAP under the full sensor configuration (LC+LC), outperforming the CoopDet3D baseline by 1.17\%, and reaches an even higher 43.38\% 3D mAP in the L+LC scenario, while maintaining 3D mAP in the range of 21.74\%–43.38\% across nine heterogeneous sensor configurations. These results, validated by our first-place finish in the CVPR 2025 DriveX challenge, establish HeCoFuse as the current state-of-the-art on TUM-Traf V2X dataset while demonstrating robust performance across diverse sensor deployments.
Code is available at \href{https://github.com/ChuhengWei/HeCoFuse}{$https://github.com/ChuhengWei/HeCoFuse$}.
\end{abstract}

\section{Introduction}

The Vehicle-to-Everything (V2X) enabled cooperative perception leverages information sharing between vehicles and roadside infrastructure to extend the perceptual horizon beyond line of sight and mitigate occlusions inherent to individual sensors~\cite{yu2022review,kim2014multivehicle,xu2022v2x}. Recent advances in V2X systems have demonstrated notable gains in detection range, accuracy, and robustness, especially under challenging conditions such as dense traffic, adverse weather, and complex urban scenes~\cite{karvat2024adver}. However, these studies predominantly assume that all nodes possess identical or similar sensor configurations, an assumption that barely holds in real-world deployments due to cost constraints, incremental hardware upgrades, and heterogeneous features in vehicles and infrastructure~\cite{xiang2023hm}.

The resulting heterogeneity creates scenarios where some nodes may carry only LiDAR, others only cameras, and yet others both sensor types\cite{wei2025integrating}. This variation poses three major challenges: First, asymmetric sensor modalities complicate feature alignment and fusion, as LiDAR and camera data differ in spatial resolution, field of view, and information density. Second, feature inconsistency arises when a modality is absent at certain nodes, rendering naive fusion strategies suboptimal. Third, system robustness must be guaranteed under partial sensor failures or when interacting with nodes of varying capabilities, requiring adaptable fusion mechanisms that degrade gracefully.

To address these challenges, we propose \textbf{HeCoFuse}, a unified cooperative perception framework tailored for heterogeneous vehicle–infrastructure (V/I) settings. HeCoFuse extracts bird’s-eye-view (BEV) features from each node via modality-specific encoders, and then performs inter-node fusion through a novel hierarchical attention-based mechanism that dynamically weighs both channel-wise and spatial features according to sensor quality. In parallel, an adaptive spatial resolution module adjusts feature-map scales based on sensor configuration to balance computational cost and information fidelity. During training, we randomly sample nine representative heterogeneous configurations, enabling the model to learn robust fusion strategies that can be generalized across all sensor combinations.

Our main contributions are threefold:
\begin{itemize}
  \item We introduce \textbf{HeCoFuse}, the first unified framework explicitly designed for cooperative perception with heterogeneous sensor configurations in V2X networks.
  \item We develop two novel mechanisms, Hierarchical Attention Fusion (HAF) and Adaptive Spatial Resolution (ASR), to dynamically integrate cross-modal features and harmonize feature scales under varying sensor availability.
  \item We conduct extensive experiments on the real-world TUMTraf-V2X dataset~\cite{zimmer2024tumtraf}, showing that HeCoFuse achieves state-of-the-art performance when trained from scratch, and maintains robust accuracy across nine heterogeneous sensor configurations.
\end{itemize}

\section{Related Work}
\subsection{Cooperative Perception Strategies}
Cooperative perception (CP) methods in V2X systems are typically categorized into three main fusion strategies—early fusion, intermediate fusion, and late fusion—which correspond to data-level fusion, feature-level fusion, and fusion of individual agents' independent outputs, respectively~\cite{huang2023v2x, han2023collaborative}. These strategies reflect different trade-offs between communication load, perception quality, and system flexibility. Early research in CP primarily explored data-level and output-level fusion approaches, while more recent work has gravitated toward intermediate fusion methods due to their balance between performance and communication efficiency.

\textbf{Early fusion} involves collaboration at the input data level, where raw sensor data from multiple agents is shared and aggregated before feature extraction. One representative method, Cooper~\cite{chen2019cooper}, integrates data from ego vehicles and other connected vehicles in a vehicular network to enhance perception capabilities. By leveraging LiDAR point clouds, Cooper fuses raw sensor data from multiple nodes at the data level. It proposes a 3D object detection framework to process aligned point cloud data, demonstrating the feasibility of performing cooperative perception through point cloud data transmission using existing vehicular network technologies. This method significantly improves detection accuracy and driving safety. To address the high communication overhead of raw data sharing, V2X-PC~\cite{liu2024v2x} introduced a novel approach by defining point clusters as message units instead of transmitting raw point cloud data, significantly reducing communication requirements while maintaining perception accuracy.

\textbf{Intermediate fusion} has emerged as the mainstream approach in CP research due to its ability to reduce communication demands while preserving essential perceptual information. Based on fusion principles, intermediate fusion techniques can be broadly classified into three categories. Traditional methods like F-Cooper~\cite{chen2019f} implement basic feature aggregation techniques such as summation and pooling operations, capitalizing on the compact nature of features to enable real-time edge computing. Attention-based methods, represented by AttentionFusion~\cite{xu2022opv2v} and V2VFormer~\cite{lin2024v2vformer}, employ various attention mechanisms to dynamically integrate multi-sensor information through weight assignment. Topology-based approaches like V2VNet~\cite{wang2020v2vnet} model multiple agents as a graph structure, using graph neural networks to facilitate information exchange. Other intermediate fusion frameworks include PillarGrid~\cite{bai2022pillargrid}, CooPercept~\cite{zhang2024coopercept}, VIMI~\cite{wang2023vimi}, HM-ViT~\cite{xiang2023hm}, CollabGAT~\cite{ahmed2024collabgat}, and V2X-ViT~\cite{xu2022v2x}, each proposing unique strategies for effective feature integration.

\textbf{Late fusion} methods integrate detection results after independent processing at each agent, typically using traditional data fusion techniques. These approaches were common during the early stages of CP development, assigning confidence levels to different modules and combining their outputs for final detection. Zhao et al.~\cite{zhao2017cooperative} proposed a method using Dempster-Shafer theory\cite{dempster1968generalization, dempster2008upper} to update detection confidence and improve lane detection accuracy, while VIPS~\cite{shi2022vips} introduced a system encoding simplified object representations for efficient post-detection matching. Other late fusion approaches include Khalifa's multi-view cooperative framework~\cite{khalifa2020novel}, Pereira's Poster platform~\cite{pereira2020poster}, and CooperFuse~\cite{zheng2024cooperfuse}, which calculates motion and scale consistency across frames for robust bounding box fusion.

Beyond these three primary categories, hybrid fusion methods combine features from different fusion stages, leveraging their respective strengths for enhanced cooperative perception. DiscoNet~\cite{li2021learning} integrates knowledge distillation to align early and intermediate fusion strategies, while ML-Cooper~\cite{xie2022soft} employs reinforcement learning to dynamically adjust information utilization under varying bandwidth constraints. Additional hybrid approaches include Arnold's dual early-late fusion method~\cite{arnold2020cooperative} and Liu's region-based fusion framework~\cite{liu2023region}, which adapt their strategies based on sensor coverage and environmental conditions.

\begin{figure*}[ht]
\centering
\includegraphics[width=0.88\linewidth]{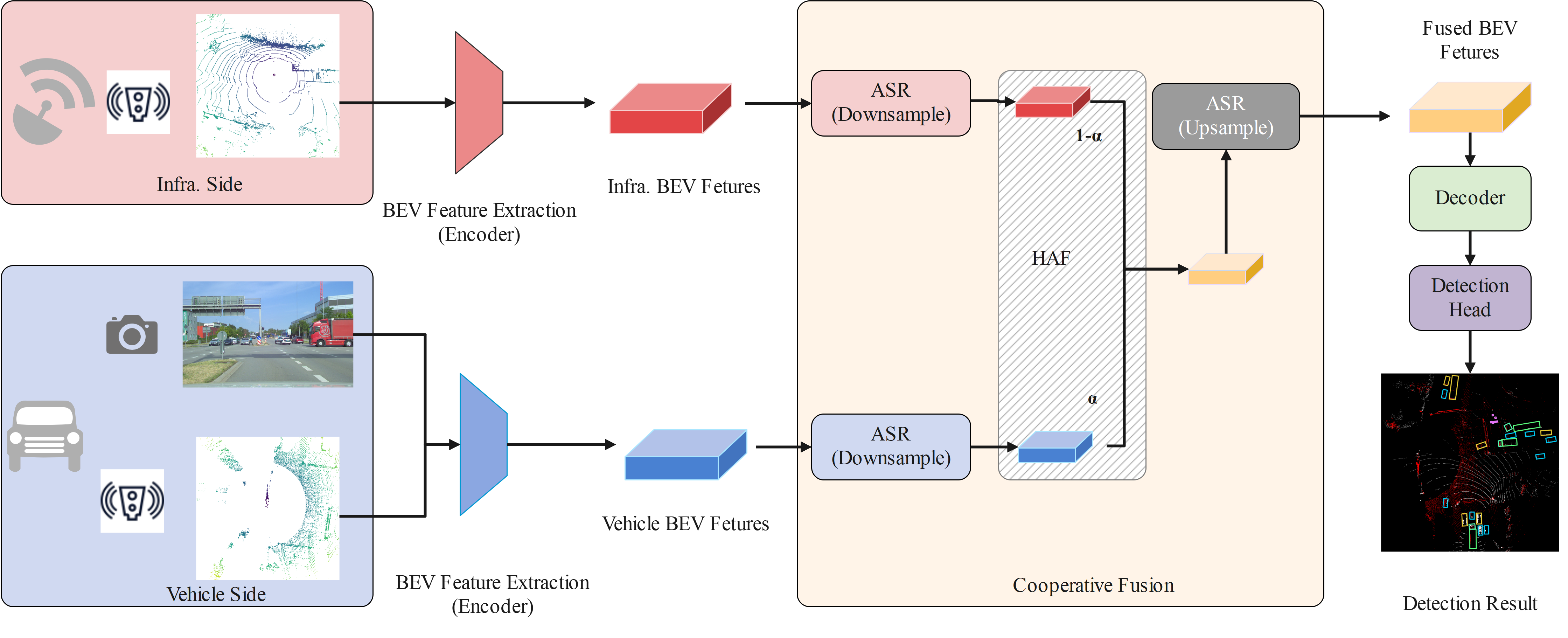}
\caption{HeCoFuse framework overview. Here, the LC+L case is used as an example (vehicle: LiDAR+Camera; infrastructure: LiDAR); for other vehicle–infrastructure sensor combinations, the per-node setups vary accordingly. The framework integrates features via Hierarchical Attention Fusion (HAF) and balances resolution through Adaptive Spatial Resolution (ASR) across different scenarios.}
\label{fig:framework}
\vspace{-1em}
\end{figure*}

\subsection{Heterogeneous Sensor Fusion}
Recent studies have begun to explore heterogeneous cooperative perception in V2X scenarios, where vehicles and infrastructure may be equipped with different sensor modalities and detection networks. HEAD~\cite{qu2024head} proposes a bandwidth-efficient cooperative perception framework that fuses the classification and regression heads of diverse 3D object detection models, enabling collaboration across vehicles equipped with different sensor types such as LiDAR and camera-based systems. By leveraging self-attention mechanisms and compact feature representations, HEAD achieves comparable perception performance to intermediate fusion while significantly reducing communication overhead. In the method proposed by Luo et al.~\cite{luo2022vehicle}, a vehicle-to-infrastructure (V2I) cooperative perception framework was proposed to address beyond visual range scenarios by combining heterogeneous sensor data. This method enhances YOLOv4~\cite{bochkovskiy2020yolov4} and CenterPoint~\cite{yin2021center} backbones through novel convolutional modules and performs multi-object post-fusion, demonstrating improvements in urban intersection settings. HM-ViT~\cite{xiang2023hm} introduces a unified hetero-modal vehicle-to-vehicle (V2V) cooperative perception framework using a heterogeneous 3D graph transformer to jointly model inter- and intra-agent interactions. Unlike prior approaches limited to homogeneous sensor configurations, HM-ViT effectively integrates LiDAR and image features across multiple dynamic agents, achieving state-of-the-art performance on OPV2V~\cite{xu2022opv2v}.

Despite the progress made by these methods, they still exhibit notable limitations. First, most of them focus on either V2V or V2I scenarios independently, rather than addressing the broader V2X cooperative setting. More importantly, their treatment of heterogeneous modalities is often limited to single-sensor configurations, such as fusing features from one camera and one LiDAR, without supporting more diverse or flexible combinations of multi-camera and multi-LiDAR setups. This restricts their scalability and effectiveness in real-world deployments where sensor configurations vary significantly across agents. To address this gap, our method aims to develop a generalized fusion framework that can accommodate arbitrary combinations of camera and LiDAR sensors across vehicles and infrastructure. By enabling more flexible hetero-modal fusion, we seek to improve perception robustness under varied deployment conditions in V2X systems.

\section{Methodology}
\subsection{Heterogeneous V2X Cooperative Perception Framework - HeCoFuse}

Real-world V2X deployments rarely feature identical sensor configurations across nodes due to cost constraints, infrastructure limitations, and hardware variations. This heterogeneity creates significant challenges for cooperative perception systems that typically assume sensor consistency. We introduce \textbf{HeCoFuse}, a unified framework designed to maintain robust perception performance across diverse sensor configurations,which framework is illustrated in Fig.~\ref{fig:framework}. Our approach addresses nine distinct sensor arrangements:

\begin{itemize}
    \item \textbf{Full sensor configuration:} LC+LC
    \item \textbf{Homogeneous single-type:} L+L, C+C
    \item \textbf{Heterogeneous single-type:} L+C, C+L
    \item \textbf{Mixed configurations:} LC+C, LC+L, C+LC, L+LC
\end{itemize}

\noindent where L denotes LiDAR and C denotes camera, and in the notation `X + Y' the symbol `+' separates the vehicle node (X) from the infrastructure node (Y). The same `X+Y' convention is used throughout this paper.

The core of HeCoFuse is a cooperative learning strategy that enables seamless operation across these heterogeneous configurations. Rather than training separate models for each sensor arrangement, we employ a unified learning approach where the network learns to extract maximum information from available sensors while maintaining consistent feature representations. This is achieved through a modular architecture with feature adaptation components that ensure dimensional compatibility between different sensor inputs. During training, we randomly sample from all nine configurations, enabling the model to learn robust feature extraction and fusion strategies that generalize across sensor combinations. This approach ensures graceful degradation when high-quality sensors are unavailable. For instance, when a vehicle with only cameras interacts with infrastructure equipped with only LiDAR sensors, the system maintains effective perception by adaptively fusing the complementary information.

\begin{figure}[ht]
\centering
\includegraphics[width=0.9\linewidth]{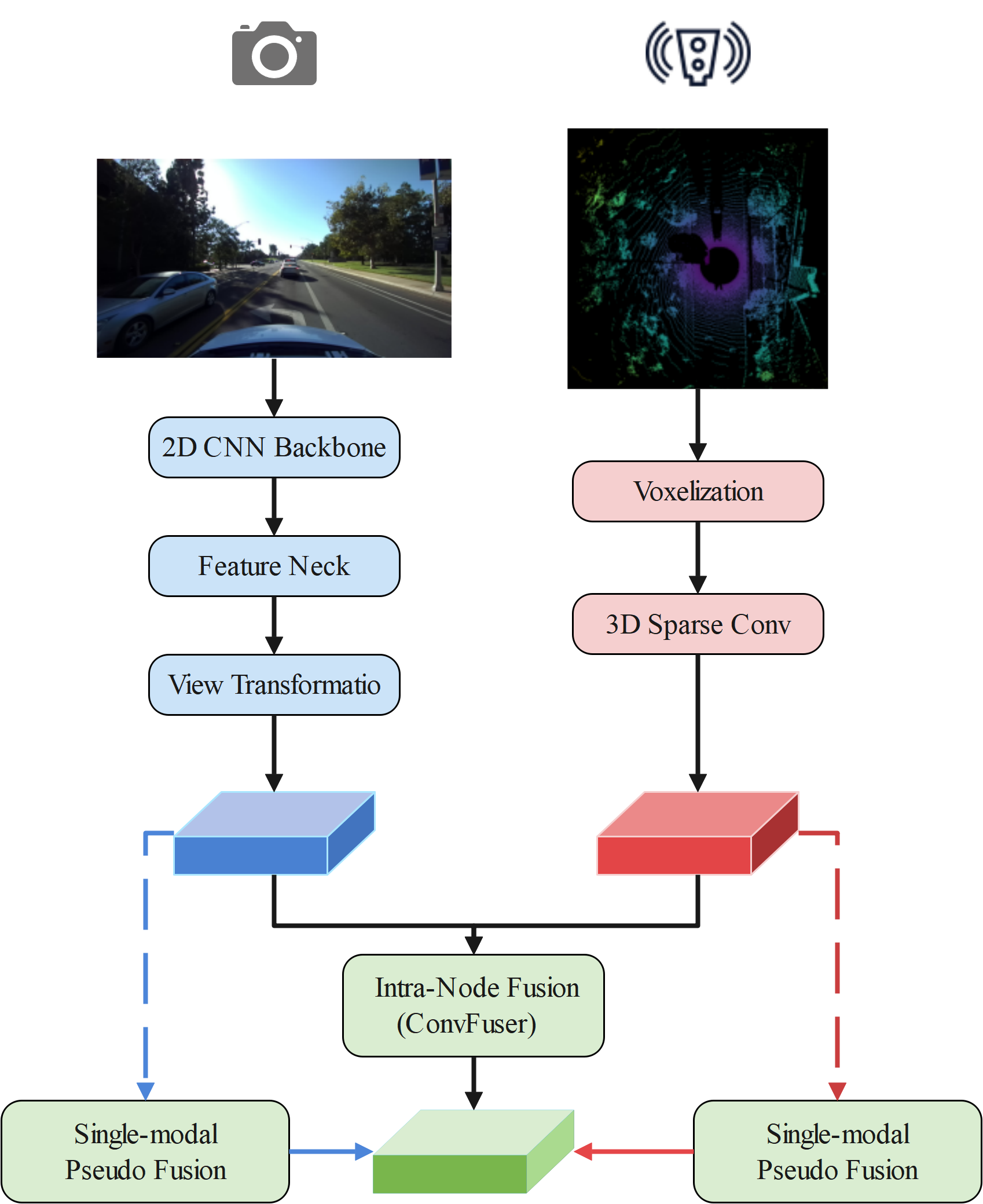}
\caption{BEV feature extraction process for heterogeneous sensors. Our framework handles various sensor configurations through specialized encoding paths for LiDAR and camera, along with intra-node fusion for multi-modal configurations and PseudoFusion for single-sensor scenarios.}
\label{fig:bev_extraction}
\end{figure}

HeCoFuse adopts a modular approach to feature extraction that adapts dynamically to available sensors at each node, as illustrated in Fig.~\ref{fig:bev_extraction}. For a given node $n \in \{vehicle, infrastructure\}$ with sensor set $S_n \subseteq \{camera, lidar\}$, features are extracted through sensor-specific pathways before fusion:
\begin{equation}
F_n = \begin{cases}
\text{Fuse}(F_n^{camera}, F_n^{lidar}), & \text{if } |S_n| = 2 \\
F_n^{s}, & \text{if } |S_n| = 1, s \in S_n
\end{cases}
\end{equation}

\subsubsection{\textbf{LiDAR Feature Encoding}}

For LiDAR processing, we transform unstructured point cloud data into Bird's Eye View (BEV) features through a multi-stage pipeline. First, our dynamic voxelization module converts raw point clouds into a structured 3D grid, adaptively handling varying point densities across different scenes. The voxelization process assigns points to voxel cells and aggregates information within each cell, significantly reducing computational complexity while preserving essential spatial information.

These voxelized features are then processed by a 3D sparse convolutional backbone that efficiently extracts spatial patterns. By operating only on non-empty voxels, the sparse convolution maintains computational efficiency while capturing detailed geometric structure. Finally, the 3D features are projected onto the ground plane to create a dense bird's-eye-view representation that serves as input to subsequent fusion stages. This approach effectively leverages the precise spatial information provided by LiDAR while remaining computationally tractable through sparse operations.

\subsubsection{\textbf{Camera Feature Encoding}}

The camera pathway transforms perspective images into BEV space through a sequential process designed to maintain rich semantic information while establishing geometric consistency. Starting with multi-view images, we apply a 2D convolutional backbone to extract features that capture appearance, texture, and object information. These initial features are enhanced through a feature neck network (e.g., FPN) that creates multi-scale representations, providing both fine-grained details and high-level semantic understanding.

The critical step in camera BEV encoding is the view transformation module, which converts perspective view features to BEV space. This involves first estimating depth distributions for each image pixel and then projecting features along these distributions into 3D space using camera-to-LiDAR transformation matrices and camera intrinsic parameters. Finally, the height dimension is collapsed to form the BEV representation. This approach addresses the inherent challenges in transforming 2D perspective views to a 3D coordinate system, enabling effective fusion with LiDAR features despite their fundamentally different sensing modalities.

\subsubsection{\textbf{BEV Fusion}}

When a node contains both LiDAR and camera sensors (e.g., in LC+LC, LC+L, LC+C configurations), we apply a BEV Fusion module to combine complementary information from both modalities:
\begin{equation}
F_n = \text{BEVFusion}(F_n^{camera}, F_n^{lidar})
\end{equation}

Our BEVFusion concatenates the features along the channel dimension and applies a convolutional layer with batch normalization and activation:

\begin{equation}
\text{BEVFusion}(F_1, F_2) = \phi_{act}(\phi_{norm}(\phi_{conv}([F_1, F_2])))
\end{equation}

This intra-node fusion leverages the complementary strengths of each modality—LiDAR provides accurate spatial information while cameras offer rich semantic features. The fused representation enhances detection performance by combining the strengths of both sensors.
\subsubsection{\textbf{Partial Sensor Configuration (LiDAR-only / Camera-only)}}

A key challenge in heterogeneous V2X systems is handling nodes with single sensor types. For these configurations, we employ PseudoFusion to ensure feature compatibility across different nodes. 

For L-only or C-only configurations, PseudoFusion employs sensor-specific feature adapters to transform features to a consistent dimensionality:
\begin{equation}
F_n^{adapted} = \phi_{adapter}^s(F_n^s), \quad s \in \{lidar, camera\}
\end{equation}

\noindent where $\phi_{adapter}^s$ is a sensor-specific 1×1 convolutional layer followed by normalization and activation. This ensures that regardless of the input sensor, the output features maintain consistent channel dimensions:
\begin{equation}
F_n^{adapted} \in \mathbb{R}^{B \times C_{out} \times H \times W}
\end{equation}

To handle the absence of a specific sensor, PseudoFusion checks sensor availability and creates appropriate feature representations:
\begin{equation}
F_n = \begin{cases}
F_n^{adapted}, & \text{sensor $s$  available} \\
\phi_{fallback}(B, C_{out}, H, W), & \text{no sensors available}
\end{cases}
\end{equation}

\noindent where $\phi_{fallback}$ creates a baseline feature representation for extreme cases where sensor data is temporarily unavailable. This approach ensures that HeCoFuse maintains robust performance across all nine possible sensor configurations by guaranteeing feature compatibility regardless of sensor availability.

\subsection{Joint Attention-based Fusion with Adaptive Resolution}

The core challenge in heterogeneous V2X fusion lies in effectively combining features from nodes with different sensor configurations while maintaining computational efficiency. As illustrated in Fig.~\ref{fig:fusion_mechanism}, we address this through a two-pronged approach: hierarchical attention fusion and adaptive spatial resolution adjustment.

\subsubsection{\textbf{Hierarchical Attention Fusion (HAF)}}

Given BEV features $F_{vehicle}$ and $F_{infra}$ from potentially asymmetric sensor sources, our channel attention mechanism first assesses which feature channels provide the most reliable information:

\begin{equation}
\alpha = \sigma(W_{channel}) \in \mathbb{R}^{1 \times C \times 1 \times 1}
\end{equation}

For heterogeneous configurations, this enables the network to adaptively weight channels from different sensor modalities. For instance, in an L+C configuration (vehicle with LiDAR, infrastructure with camera), $\alpha$ will preferentially weight LiDAR-dominated channels for distance estimation while favoring camera-dominated channels for appearance features:
\begin{equation}
F_{channel} = \alpha \odot F_{vehicle} + (1-\alpha) \odot F_{infra}
\end{equation}

\begin{figure}[h]
\centering
\includegraphics[width=0.85\linewidth]{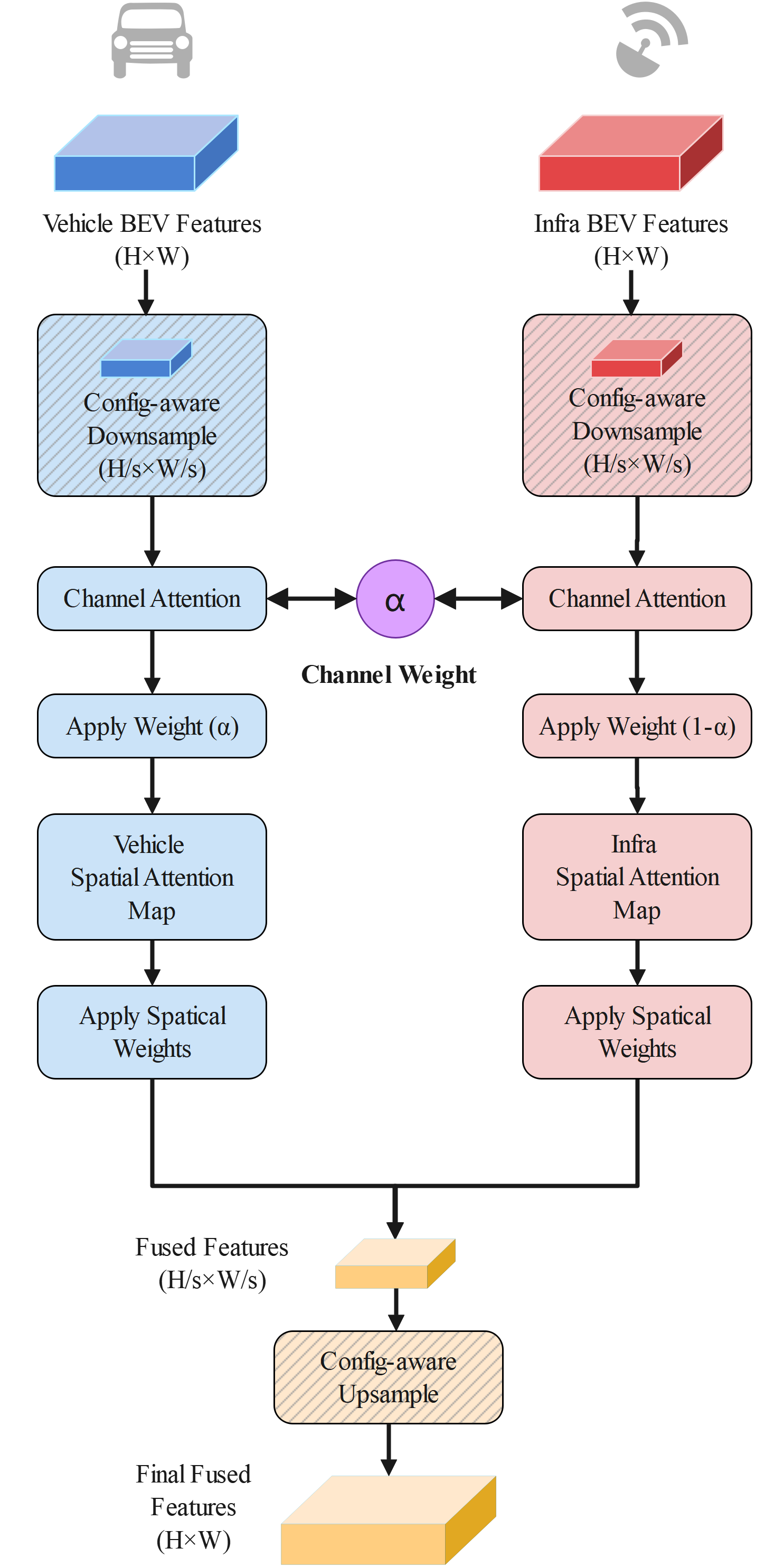}
\caption{Architecture of inter-node feature fusion combining Hierarchical Attention Fusion (HAF) and Adaptive Spatial Resolution (ASR) mechanisms. The shaded regions denote the ASR modules.}
\label{fig:fusion_mechanism}
\end{figure}

As shown in the central portion of Fig.~\ref{fig:fusion_mechanism}, our spatial attention further refines this fusion by addressing modality-specific strengths in different spatial regions. For nodes with varying sensor configurations, spatial attention maps $A_n$ highlight regions where each node's particular sensors provide the most reliable information:
\begin{equation}
A_n = \sigma(\phi_{conv2}(\phi_{conv1}(F_n))) \in \mathbb{R}^{B \times 1 \times H \times W}
\end{equation}

In heterogeneous settings, these attention maps become particularly important—when a vehicle with only LiDAR interacts with infrastructure with only cameras, spatial attention emphasizes LiDAR's strengths in accurate distance measurements for distant objects while prioritizing camera-based detections in visually complex regions:
\begin{align}
F_{fused} =& (F_{vehicle} \odot \alpha) \odot A_{vehicle} \\ \nonumber
&+ (F_{infra} \odot (1-\alpha)) \odot A_{infra}
\end{align}

\noindent where $\odot$ denotes the Hadamard (element-wise) product, allowing for fine-grained weighting of each feature element independently.

\subsubsection{\textbf{Adaptive Spatial Resolution (ASR)}}

Different sensor configurations produce BEV features with varying characteristics, requiring different computational resources. As depicted in the upper and lower portions of Fig.~\ref{fig:fusion_mechanism}, our approach incorporates adaptive resolution adjustment that works in concert with attention fusion.

For heterogeneous configurations, we first apply configuration-aware downsampling:
\begin{align}
F_{n,down} &= \phi_{downsample}(F_n, s_n)\\ \nonumber
&= \text{AdaptiveAvgPool2d}(F_n, H/s_n, W/s_n),
\end{align}

\noindent where $s_n$ is determined based on the specific sensor configuration at node $n$:
\begin{equation}
s_n = \begin{cases}
s_{high}, & \text{if node $n$ has camera-only} \\
s_{low}, & \text{if node $n$ has LiDAR-only} \\
s_{medium}, & \text{if node $n$ has both sensors}
\end{cases}
\end{equation}

Hierarchical attention fusion is then applied to these downsampled features. After fusion, we apply configuration-aware upsampling that prioritizes restoration of detailed information from the most precise sensor configuration:
\begin{equation}
F_{final} = \phi_{upsample}(F_{fused}, \min(s_{vehicle}, s_{infra}))
\end{equation}

This integrated approach, combining hierarchical attention fusion with adaptive resolution adjustment, enables HeCoFuse to effectively bridge the gap between different sensor configurations while reducing computational overhead by 30-45\% in heterogeneous configurations.

\begin{table*}[ht]
\centering
\caption{Quantitative comparison of HeCoFuse with baseline and challenge methods across different sensor configurations.}
\begin{threeparttable} 
\begin{tabular}{lccccccccc}
\hline
\multirow{3}{*}{Method} & \multirow{2}{*}{Type} & \multicolumn{2}{c}{Config} & \multirow{2}{*}{Precision } & \multirow{2}{*}{Recall} & \multirow{2}{*}{3D IoU } & \multirow{2}{*}{Pos. RMSE } & \multirow{2}{*}{Rot. RMSE } & \multirow{2}{*}{3D mAP}\\
\cmidrule(lr){3-4} 
 & &  Vehicle & Infra. & ($\uparrow$) & ($\uparrow$) & ($\uparrow$) & ($\downarrow$) & ($\downarrow$) &($\uparrow$) \\
\hline
CoopDet3D $^{*}$ & & LC&LC & 42.34 & 42.86 & 0.36  & 0.48 &  \textbf{0.03} & 42.05\\
BUPTMM $^{\dagger}$ & & LC & LC & 43.63 &\textbf{44.55}& \textbf{0.38} &\textbf{0.45}&0.04 &43.35 \\
CV123 $^{\dagger}$ & & LC & LC  & 43.53&44.32&\textbf{0.38}&\textbf{0.45}&\textbf{0.03}&43.25\\
TEAM12138 $^{\dagger}$ & & LC & LC&43.52&44.32&\textbf{0.38}&\textbf{0.45}&\textbf{0.03}&43.25 \\
KaAI $^{\dagger}$ & & LC & LC &43.51&44.35&\textbf{0.38}&\textbf{0.45}&\textbf{0.03}&43.22\\
\hline
HeCoFuse & \multirow{1}{*}{Full sensor configuration} & LC&LC & 43.52 & 44.34 &\textbf{0.38} & \textbf{0.45} & \textbf{0.03} & 43.22 \\

\hline
HeCoFuse & \multirow{2}{*}{Homogeneous single-type} & L&L & 42.50 & 43.52 & 0.33 & 0.48 & 0.06 & 42.10 \\
HeCoFuse & & C&C & 22.30 & 17.33 & 0.18 & 0.65 & 0.15 & 21.74\\
\hline
HeCoFuse & \multirow{2}{*}{Heterogeneous single-type}& L&C & 30.52 & 26.64 & 0.21 & 0.65 & 0.13 & 30.04 \\
HeCoFuse & & C&L & 33.22 & 24.72 & 0.21 & 0.51 & 0.11 & 32.68 \\
\hline
HeCoFuse & \multirow{4}{*}{Mixed configurations:} & LC&C &31.45 & 27.37 & 0.20 & 0.72 & 0.10  & 30.85 \\
HeCoFuse & & LC&L & 39.64 & 38.53 & 0.34 & 0.52 & 0.05 & 39.17 \\
HeCoFuse & & C&LC & 35.27 & 33.08 & 0.26 & 0.57 & 0.11 & 34.76 \\
HeCoFuse & & L&LC & \textbf{43.73} & 42.91 & 0.33 & 0.50 & 0.04 & \textbf{43.38} \\
\hline
\end{tabular}
\begin{tablenotes}
    \item The \textbf{bolded values} indicate the best-performing model for each metric.
    \item * The results reported in \cite{zimmer2024tumtraf} are obtained using pre-trained Camera and LiDAR models, while in our reproduction, the training is performed from scratch.
    \item $\dagger$ These methods represent the other top-5 performing algorithms from the CVPR DriveX Workshop TUMTraf-V2X Challenge~\cite{drivex2025challenge}, excluding HeCoFuse.
\vspace{-1em}
\end{tablenotes}
\end{threeparttable}
\label{tab}
\vspace{-1em}
\end{table*}

\section{Experimental Evaluations}
\subsection{Dataset Selection}
Both simulated and real-world datasets offer distinct advantages for cooperative perception research. Simulated datasets like V2XSet~\cite{xu2022v2x} and OPV2V~\cite{xu2022opv2v} provide controlled environments, while real-world datasets such as DAIR-V2X~\cite{yu2022dair}, V2X-Seq~\cite{yu2023v2x}, and V2XReal~\cite{xiang2024v2x} capture authentic traffic dynamics and sensor noise. For our heterogeneous cooperative perception framework, we require a dataset meeting several criteria: real-world data collection, multi-modal sensing with both LiDAR and camera streams at each node, consistent sensor types across nodes, and precise calibration between vehicle and infrastructure. The dataset should also feature diverse traffic scenarios with sufficient annotations to evaluate detection performance.

We selected the TUMTraf-V2X dataset~\cite{zimmer2024tumtraf} for our experiments due to its advantages for heterogeneous cooperative perception research. Collected at an urban intersection in Munich, it includes synchronized data from vehicle and infrastructure perspectives with 2,000 camera images, 5,000 LiDAR point clouds, and 29,000 annotated 3D bounding boxes across 8 object categories. Both nodes feature mechanical LiDAR sensors and cameras with similar specifications, enabling simulation of various heterogeneous configurations while maintaining a consistent baseline for evaluation.

\begin{figure*}[t]
\centering
\includegraphics[width=\linewidth]{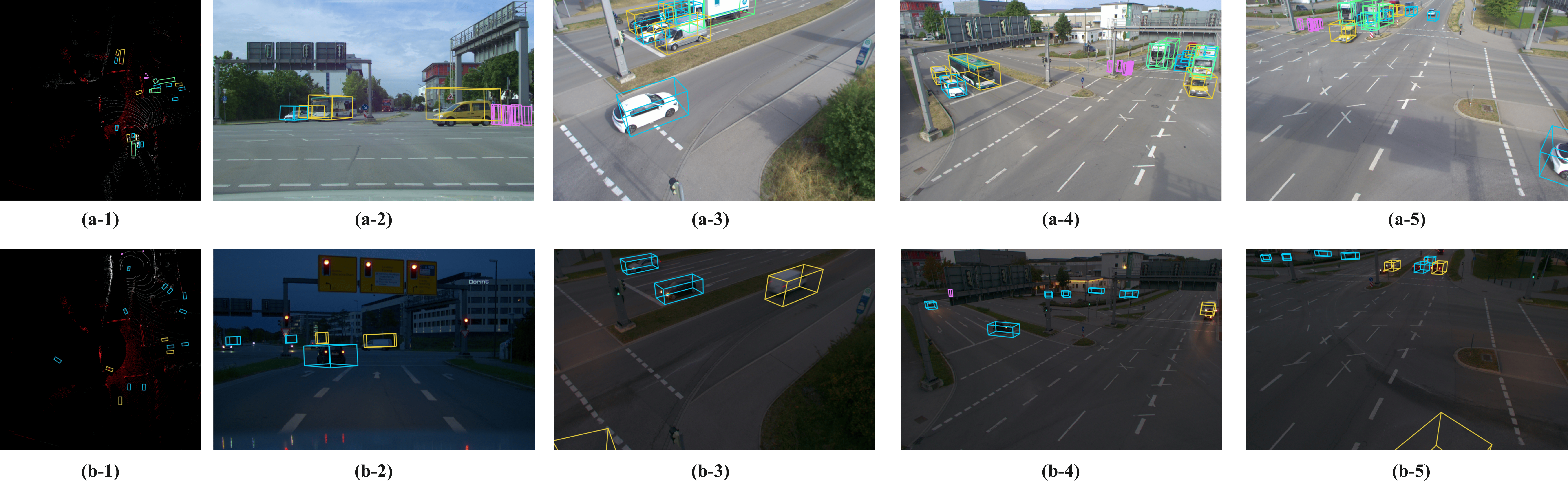}
\caption{Qualitative visualization of HeCoFuse detection results in the full sensor configuration (LC+LC) on the TUMTraf-V2X test set. The figure illustrates detection results in both daytime (first row) and nighttime (second row) conditions. (a-1, b-1) BEV perspective showing fused LiDAR point clouds with detection results from both vehicle and infrastructure; (a-2, b-2) vehicle-mounted camera view with corresponding 2D projections of 3D detections; (a-3, a-4, a-5, b-3, b-4, b-5) three different infrastructure camera views with projected detection results.}
\label{fig-results}
\end{figure*}

\subsection{Quantitative Analysis}
We trained our HeCoFuse framework from scratch on the TUMTraf-V2X mini dataset (half of the complete dataset) for 20 epochs using a single GeForce RTX 4090 GPU. For both vehicle and infrastructure nodes, we employed YOLOv8~\cite{Jocher_Ultralytics_YOLO_2023} as the backbone encoder for camera data processing, while PointPillars~\cite{lang2019pointpillars} was utilized as the encoder for LiDAR point cloud processing.
To comprehensively evaluate our method, we adopted multiple metrics including Precision, Recall, 3D IoU, Position RMSE, Rotation RMSE, and 3D mAP across various object categories. Table~\ref{tab} presents the performance comparison against the official TUMTraf-V2X baseline method, CoopDet3D~\cite{zimmer2024tumtraf}. These metrics evaluate different aspects of detection performance: Precision and Recall measure detection accuracy, 3D IoU quantifies the spatial overlap between predicted and ground truth bounding boxes, Position and Rotation RMSE assess localization accuracy, while 3D mAP provides a comprehensive measure of detection performance across varying confidence thresholds.

Our HeCoFuse method achieved first place in the CVPR 2025 DriveX Workshop TUMTraf-V2X Challenge~\cite{drivex2025challenge} based on the 3D mAP metric with the L+LC configuration (vehicle with LiDAR only, infrastructure with both LiDAR and camera), reaching 43.38\% 3D mAP. Across all six evaluation metrics, HeCoFuse achieves the best or tied-for-best performance on five metrics, demonstrating the superiority of our approach. Under the full sensor configuration (LC+LC), HeCoFuse achieves 43.22\% 3D mAP, representing a 1.17\% improvement over the baseline's 42.05\%. Notably, the L+LC configuration delivers the highest overall performance with 43.38\% 3D mAP and superior precision (43.73\%), highlighting the effectiveness of our hierarchical attention fusion mechanism in heterogeneous settings.

The results reveal several important insights about heterogeneous sensor fusion. First, LiDAR sensors contribute significantly more to detection performance than cameras in our framework. For instance, comparing L+L (42.10\% mAP) with C+C (21.74\% mAP) configurations shows a substantial 20.36\% performance gap, indicating LiDAR's superior spatial accuracy for 3D object detection. Second, infrastructure sensors play a pivotal role in overall system performance due to their wider field of view and elevated mounting position. This is evident when comparing L+LC (43.37\% mAP) with LC+L (39.17\% mAP), where the former achieves 4.20\% higher mAP despite using fewer total sensors. Interestingly, some metrics for the L+LC configuration even surpass those of the fully-equipped LC+LC case; this may be because the roadside camera already provides sufficiently rich perspectives, and the vehicle’s additional camera, limited by its mounting viewpoint, adds little extra benefit.

These findings validate our adaptive fusion approach's capability to dynamically leverage the most reliable information sources across heterogeneous configurations, maintaining robust performance even with sensor asymmetry. This represents a critical capability for real-world V2X deployments where sensor configurations often vary widely.

\subsection{Qualitative Analysis}

To provide a comprehensive understanding of HeCoFuse's performance in real-world scenarios, we present qualitative visualization results using the full sensor configuration (LC+LC) on the TUMTraf-V2X test set, as shown in Fig.~\ref{fig-results}. The visualization demonstrates our system's detection capabilities from multiple perspectives: the BEV representation of fused LiDAR point clouds, the vehicle's camera view, and three different infrastructure camera viewpoints. We specifically selected examples from both daytime and nighttime conditions to illustrate the robustness of our approach across varying lighting conditions.

Our framework successfully detects multiple object categories including passenger vehicles, buses, trains, and pedestrians across different viewing angles and environmental conditions. Notably, the detection performance remains consistent in challenging nighttime scenarios (b-1 to b-5), where camera-only methods typically struggle due to poor lighting. This robustness can be attributed to our hierarchical attention fusion mechanism, which dynamically adjusts the contribution of each sensor modality based on its reliability in different environmental conditions. For instance, in nighttime scenarios, the fusion mechanism automatically places greater emphasis on LiDAR features, which are less affected by lighting variations, while still leveraging the complementary semantic information from cameras. The multi-perspective visualization further demonstrates how HeCoFuse effectively integrates information from spatially distributed sensors, addressing occlusion issues and extending the perceptual range beyond what would be possible with a single sensor node. These qualitative results reinforce our quantitative findings and highlight the practical advantages of heterogeneous sensor fusion for robust cooperative perception in complex urban environments.

\section{Conclusion}

In this paper, we introduce HeCoFuse, a unified cross-modal cooperative perception framework specifically designed to address the challenges of heterogeneous sensor configurations in vehicle-infrastructure cooperative perception systems. Our approach bridges the gap between idealized research assumptions and real-world deployments where vehicles and infrastructure possess varying sensor types and capabilities. HeCoFuse incorporates several key innovations: a hierarchical attention-based fusion mechanism that dynamically integrates features based on their quality across different modalities and nodes, an adaptive spatial resolution adjustment module that balances computational efficiency with information preservation, and a cooperative learning strategy that enables consistent performance across nine different sensor configurations. Through extensive experiments on the real-world TUMTraf-V2X dataset, we demonstrate that HeCoFuse achieves state-of-the-art performance in fully equipped settings while maintaining robust accuracy in heterogeneous scenarios, particularly when infrastructure nodes are equipped with multiple sensor modalities. Notably, our framework shows exceptional resilience in asymmetric configurations common in practical deployments, where some nodes may only have partial sensing capabilities.

Several promising directions remain for future research. First, extending the validation to scenarios with more than two nodes would further demonstrate the scalability of our approach in complex urban environments with multiple vehicles and infrastructure units. Second, challenging weather conditions such as rain, fog and snow represent critical test cases for heterogeneous fusion approaches, as different sensor modalities exhibit complementary strengths and weaknesses under adverse conditions. Such conditions place greater demands on cooperative perception systems and would further validate the adaptive fusion capabilities of our framework. Finally, expanding the camera configurations to incorporate additional viewpoints on the vehicle side could help address the performance gap between LiDAR and camera-based detection, potentially enabling more balanced multi-modal fusion even when LiDAR sensors are unavailable. This would be particularly valuable for low-cost cooperative perception systems where camera sensors might be the primary sensing modality available on some nodes.

\bibliographystyle{IEEEtran}

\bibliography{main.bib}

\end{document}